\documentclass[10pt,a4paper]{article}

\usepackage[utf8]{inputenc}
\usepackage[T1]{fontenc}
\usepackage{amsmath,amssymb,amsfonts}
\usepackage{graphicx}
\usepackage{hyperref}
\usepackage{algorithm}
\usepackage{algpseudocode}
\usepackage{multicol}
\usepackage{geometry}
\usepackage{cite}
\usepackage{authblk}

\title{\Large\textbf{PixelBytes: Catching Unified Representation for Multimodal Generation}}
\author{\large Fabien Furfaro\thanks{\texttt{fabien.furfaro@capgemini.com}}}
\date{\large 2024}

\begin{document}

\maketitle

\begin{abstract}
This report presents PixelBytes, an approach for unified multimodal representation learning. Drawing inspiration from sequence models like Image Transformers, PixelCNN, and Mamba-Bytes, we explore integrating text, audio, action-state, and pixelated images (sprites) into a cohesive representation. We conducted experiments on a PixelBytes Pokémon dataset and an Optimal-Control dataset. Our investigation covered various model architectures, including Recurrent Neural Networks (RNNs), State Space Models (SSMs), and Attention-based models, with a focus on bidirectional processing and our PxBy embedding technique. We evaluated models based on data reduction strategies and autoregressive learning, specifically examining Long Short-Term Memory (LSTM) networks in predictive and autoregressive modes. Our results indicate that autoregressive models perform better than predictive models in this context. Additionally, we found that diffusion models can be applied to control problems and parallelized generation. PixelBytes aims to contribute to the development of foundation models for multimodal data processing and generation. The project's code, models, and datasets are available online \cite{furfaro2024pixelbytes_project}.
\end{abstract}

\begin{figure}[ht]
\centering
\includegraphics[width=0.8\linewidth]{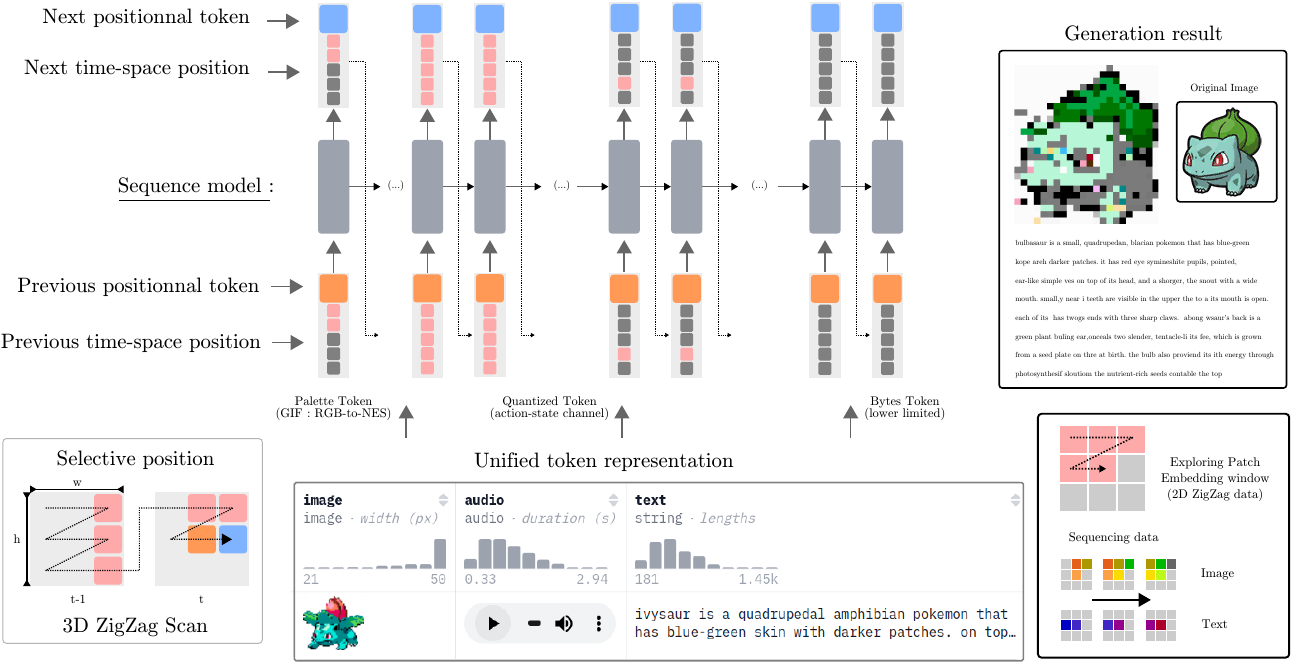}
\caption{Overview of the PixelBytes approach: (Left) Scanning process reads different modalities at specific positions. (Center) Example of our dataset and autoregressive generation across modalities. (Right) Generation example, with PxBy embedding window displayed below. Note: The model currently shows some inaccuracies in image generation and may produce incorrect words, indicating areas for future improvement.}
\label{fig:approach_overview}
\end{figure}

\section{Introduction}

Recent advancements in artificial intelligence have led to increasingly generalist models, not by combining multiple specialized components (like Gato from DeepMind \cite{reed2022generalist}), but by assigning simple tasks to models where emergent properties—complex behaviors arising from simpler underlying rules—appear. This is exemplified by generative language models such as GPT \cite{brown2020language}. However, these models are constrained by their focus on language alone, failing to capture the full complexity of multimodal understanding \cite{huang2023language}. To address this limitation, researchers have explored integrating Large Language Models (LLMs) with other modalities \cite{liu2024visual}. However, this approach often results in specialized model combinations without fostering new emergent properties. We propose "PixelBytes", a novel approach enabling unified training across modalities by representing diverse inputs in a single, cohesive format.

Multimodal sequence generation, which involves creating coherent outputs combining various data types such as text, images, and numerical sequences, presents a significant challenge in artificial intelligence \cite{baltruvsaitis2018multimodal}. While models like GPT have excelled in text generation \cite{brown2020language}, there is a growing need for unified approaches that can seamlessly handle diverse data types. Building on these findings, PixelBytes addresses the challenge of unified text, audio, and image generation by proposing a model capable of producing coherent mixed sequences of text, audio, and images. Our approach is inspired by state-of-the-art sequence models, including Image Transformers \cite{parmar2018image}, PixelCNN \cite{van2016conditional}, and recent developments in byte generation by Mamba architectures \cite{wang2024mambabyte}.

Our research investigates various architectures including Recurrent Neural Networks (RNNs), State Space Models (SSMs) \cite{dao2024transformers}, and Attention-based models, examining the effectiveness of bidirectional processing \cite{zhu2024vision, liang2024bi, schuster1997bidirectional}, innovative embedding techniques (particularly PxBy embedding, which integrates pixel and byte-level representations), the influence of convolutional layers, and the effects of model depth, input dimensionality, and size.

Our experiments, conducted on a specialized PixelBytes Pokémon dataset, suggest that autoregressive models with balanced data reduction strategies perform better than predictive models in terms of accuracy and loss. We initially explored bidirectional sequence models with PxBy embedding and convolutional layers, but our final results focus on Long Short-Term Memory (LSTM) networks in both predictive and autoregressive modes. Additionally, we tested our approach on control problems using diffusion models. This paper presents our method for constructing unified sequence data from text, audio, action-state, and pixelated images (sprites), along with our experimental findings and analysis. Through a flexible approach to multimodal modeling, PixelBytes seeks to contribute to the development of foundation models capable of processing and generating multimodal data.

\section{Exploration for a Unified Representation}

\subsection{Hypothesis Testing Framework}

The quest for a unified data representation across different modalities presents significant challenges. Text data typically exhibits a one-dimensional dependency on preceding words with discrete values\cite{bengio2003neural}. Audio signals have a temporal dimension with continuous values\cite{van2016wavenet}. Action-state representations in robotics can be similar to audio but with correlations between channels\cite{botteghi2021low}. Images and animations combine spatial and temporal dimensions across discrete RGB channels\cite{krizhevsky2012imagenet}. Given these diverse characteristics, many researchers have focused on combining separate embedding models into a single framework, as seen in projects like ImageBind\cite{girdhar2023imagebind} and RT-2\cite{zitkovich2023rt}, rather than seeking a truly unified data representation. However, to build a model that comprehensively understands different modalities without intermediate alignment steps, exploring a unified data representation becomes necessary. In this section, we will examine several hypotheses:

\begin{itemize}
    \item Can we quantize data such that each element becomes a token?\cite{zhan2022auto}
    \item Is predicting only the next value sufficient for a sequence model to learn effectively?
    \item For dimensions higher than one, is applying a convolutional filter necessary?
    \item For space-time dependency, what is the importance of bidirectionality in models?
\end{itemize}

Through these investigations, we aim to explore a method of representing data that could enable models to understand various types of modalities.

\subsection{Conceptual Multimodal Embedding}

\subsubsection{Dataset Construction}

To evaluate our hypotheses on unified representation, we required a dataset combining visual and textual data suitable for byte-level processing. Image captioning datasets proved inadequate due to limited text content and challenges in interpreting pixelated versions of high-resolution images. Consequently, we created a specialized Pokémon dataset, offering pixelated designs and rich descriptive text. Data was collected by web scraping Pokémon miniatures and descriptions from Pokepedia using Beautiful Soup \cite{richardson2007beautiful}, maintaining a 2/3 text to 1/3 image ratio. For image processing, we utilized a 55-color palette inspired by the NES, creating tokens for various color combinations. This approach enabled us to represent visual information in a format compatible with our tokenizer.

\sloppy

To manage transitions between text and image tokens, we developed a 2D input sequence method utilizing a 3x3 context window around each token with a 2D zigzag scheme (Figure \ref{fig:approach_overview}). Special tokens denote transitions between text and images, with padding added to maintain consistent context sizes. The padding value is 0, line transition value is 1, and modality transition value is 2. For text, only preceding tokens are included in the context windows. Employing OpenCV and scikit-image \cite{bradski2000opencv, van2014scikit} for image quantization and pixelization, we adjusted all entries to have 113 indices, balancing text and image tokens. The resulting dataset, combining text and pixelated images, is available on the Hugging Face Datasets Hub \cite{furfaro2024pixelbytes_project} for reproducibility. It includes a "pixelbyte" column for this specific data representation.

\subsubsection{Embedding Techniques}

Our exploration of unified representation begins with integrating image-text data for sequence generation. We developed a tokenizer that processes the "pixelbyte" column. This is paired with an embedding technique called PxByEmbed, which creates a unified representation for pixel and byte data in a single space. PxByEmbed is designed to test our hypotheses about the effectiveness of convolutional filters for higher-dimensional data. It uses a learned embedding matrix to map each token (text or image) to a vector space, while maintaining spatial relationships for image tokens. PxByEmbed incorporates a simple convolutional layer and an adaptive mixing mechanism. This design allows us to investigate whether predicting only the next value (with or without only previous value) is sufficient for effective learning in a sequence model.

\begin{algorithm}[ht]
\footnotesize
\caption{PxByEmbed: Multimodal Embedding Algorithm (k=3)
\newline
\textbf{Input:} $V$: vocabulary size, $D$: embedding dimension
\newline
\textbf{Output:} Embedded representation $\mathbf{E} \in \mathbb{R}^{B \times L \times D}$
\newline
\textbf{Note:} $\mathbf{X}_{emb} \in \mathbb{R}^{B \cdot L \times E_{int} \times k \times k}$, 
$\mathbf{X}_{flat} \in \mathbb{R}^{B \cdot L \times E_{int}k^2}$, 
$\mathbf{X}_{proj} \in \mathbb{R}^{B \cdot L \times D}$
}
\begin{algorithmic}[0]
\State \textbf{Initialize:}
\State $k \gets 3$
\State $E_{int} \gets \max(9, \lfloor D / k^2 \rfloor)$
\State $\mathbf{\alpha} \in \mathbb{R}^{1 \times 1 \times k \times k}$ 
\State $\mathbf{W}_{emb} \in \mathbb{R}^{V \times E_{int}}$ 
\State $\mathbf{W}_{proj} \in \mathbb{R}^{E_{int}k^2 \times D}$ 
\State $\mathbf{W}_{patch} \in \mathbb{R}^{E_{int} \times E_{int} \times k \times k}$ 

\Function{PxByEmbed}{$\mathbf{X} \in \mathbb{Z}^{B \times L \times k \times k}$}
    \State $\mathbf{X}_{emb} \gets \text{Permute}(\text{Embed}(\mathbf{X}, \mathbf{W}_{emb}), [0, 3, 1, 2])$ 
    
    \State $\mathbf{X}_{patch} \gets \text{Conv2D}(\mathbf{X}_{emb}, \mathbf{W}_{patch}, \text{padding}=1)$
    \State $\mathbf{X}_{combined} \gets \sigma(\mathbf{\alpha}) \odot \mathbf{X}_{emb} + (1 - \sigma(\mathbf{\alpha})) \odot \mathbf{X}_{patch}$
    
    \State $\mathbf{X}_{flat} \gets \text{Flatten}(\mathbf{X}_{combined})$ 
    \State $\mathbf{X}_{proj} \gets \mathbf{X}_{flat}\mathbf{W}_{proj}$ 
    \State $\mathbf{E} \gets \text{LayerNorm}(\mathbf{X}_{proj})$
    \State $\mathbf{E} \gets \text{Reshape}(\mathbf{E}, [B, L, D])$
    \State \Return $\mathbf{E}$
\EndFunction
\end{algorithmic}
\end{algorithm}

The algorithm processes input sequences in 3x3 patches, embedding each element and then applying a learnable convolution. The embedding pad value is set to 0 to avoid influencing the training of subsequent tokens. These two representations are then combined using a learned parameter, allowing the model to adaptively balance between local and global information.

\subsection{Model Architectures Evaluated}

We evaluated three compact model architectures: a Recurrent Neural Network (RNN) using Long Short-Term Memory (LSTM) units \cite{hochreiter1997long}, a Transformer \cite{vaswani2017attention}, and a State Space Model (SSM) based on Mamba \cite{gu2023mamba}. For both the RNN and Mamba architectures, we compared variants with and without a bidirectional first layer. Each model was constrained to fewer than 100,000 parameters and adapted to process our dataset of pixel data and bytecode sequences. The models were trained on Kaggle using T4 GPUs, with a batch size of 32, sequence length of 256, and learning rate of 0.001 for 200 epochs. The trained models are available on the Hugging Face Model Hub \cite{furfaro2024pixelbytes_project}.

\begin{figure}[ht]
\centering
\includegraphics[width=\textwidth]{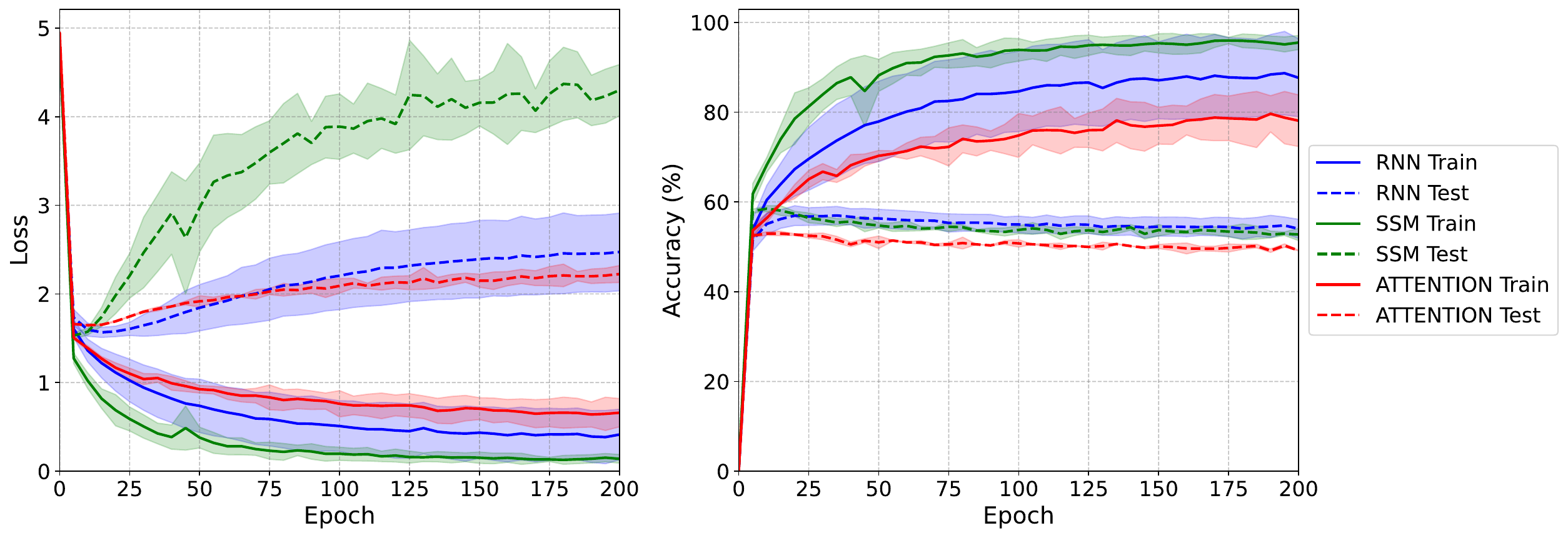}
\caption{Training and validation metrics for RNN, Transformer, and SSM models over 200 epochs.}
\label{fig:training_results}
\end{figure}

\subsection{Comparative Analysis}

Figure \ref{fig:training_results} illustrates the training and validation metrics for our three model types. The SSM achieved the best scores for loss and accuracy but exhibited signs of overfitting. The RNN demonstrated more balanced performance, suggesting better generalization to unseen data. The Transformer had the lowest performance, possibly due to the absence of positional encoding. The strong performance of the SSM supports the potential of unified representation for pixel and byte data. However, its tendency to overfit suggests that predicting only the next value might not be sufficient for effective learning in all cases. The balanced performance of the RNN indicates that simpler architectures can still be effective for our task. The Transformer's lower performance suggests that complex attention mechanisms may not always be necessary or beneficial for this type of data.

\subsubsection{Generation Evaluation Metrics}

To assess the effectiveness of our approach and various model architectures, we evaluated their generation capabilities. We tested State Space Models (SSM), Attention models (Att), and Recurrent Neural Networks (RNN) in generating 32 consecutive sequences. Our evaluation used three metrics: Hamming Distance \cite{hamming1950error}, Cosine Similarity, and BLEU Score \cite{papineni2002bleu}.

\begin{table}[ht]
\centering
\small
\begin{tabular}{cccccccccc}
\hline
Type & Dir. & Emb. & Conv. & In Emb & Hidden State & Depth & Hamming & Cosine & BLEU \\
\hline
SSM & Bi & PxBy & Y & 81 & 64 & 1 & 0.170 $\pm$ 0.086 & 0.883 $\pm$ 0.105 & 0.753 $\pm$ 0.115 \\
SSM & Bi & PxBy & Y & 81 & 64 & 2 & 0.158 $\pm$ 0.074 & 0.896 $\pm$ 0.095 & 0.771 $\pm$ 0.097 \\
SSM & Uni & PxBy & Y & 81 & 64 & 2 & 0.166 $\pm$ 0.081 & 0.886 $\pm$ 0.102 & 0.760 $\pm$ 0.106 \\
Att & - & PxBy & Y & 81 & 64 & 1 & 0.157 $\pm$ 0.064 & 0.887 $\pm$ 0.103 & 0.765 $\pm$ 0.088 \\
Att & - & PxBy & Y & 81 & 64 & 2 & 0.159 $\pm$ 0.066 & 0.887 $\pm$ 0.103 & 0.760 $\pm$ 0.092 \\
RNN & Bi & Center & N & 81 & 64 & 2 & 0.185 $\pm$ 0.074 & 0.888 $\pm$ 0.083 & 0.750 $\pm$ 0.093 \\
RNN & Bi & PxBy & Y & 81 & 64 & 2 & 0.153 $\pm$ 0.061 & 0.902 $\pm$ 0.090 & 0.777 $\pm$ 0.083 \\
RNN & Bi & PxBy & Y & 162 & 64 & 2 & 0.152 $\pm$ 0.062 & 0.905 $\pm$ 0.089 & 0.778 $\pm$ 0.084 \\
RNN & Bi & PxBy & Y & 36 & 64 & 2 & 0.152 $\pm$ 0.061 & 0.904 $\pm$ 0.090 & 0.778 $\pm$ 0.083 \\
RNN & Bi & PxBy & Y & 81 & 128 & 2 & 0.153 $\pm$ 0.063 & 0.903 $\pm$ 0.091 & 0.776 $\pm$ 0.086 \\
RNN & Bi & PxBy & Y & 81 & 32 & 2 & \textbf{0.149} $\pm$ 0.062 & 0.899 $\pm$ 0.095 & \textbf{0.785} $\pm$ 0.082 \\
RNN & Bi & PxBy & Y & 81 & 64 & 1 & 0.149 $\pm$ 0.062 & 0.897 $\pm$ 0.096 & 0.780 $\pm$ 0.085 \\
RNN & Bi & PxBy & Y & 81 & 64 & 3 & 0.153 $\pm$ 0.063 & \textbf{0.906} $\pm$ 0.087 & 0.776 $\pm$ 0.086 \\
RNN & Bi & PxBy & N & 81 & 64 & 2 & 0.151 $\pm$ 0.062 & 0.903 $\pm$ 0.090 & 0.779 $\pm$ 0.084 \\
RNN & Uni & PxBy & Y & 81 & 64 & 2 & 0.153 $\pm$ 0.064 & 0.904 $\pm$ 0.088 & 0.777 $\pm$ 0.087 \\
\hline
\end{tabular}
\caption{Comparison of model characteristics and performance (mean $\pm$ std)}
\label{tab:model_comparison}
\end{table}

The results reveal variations across different model configurations. Among RNN models with PxBy embedding, we observed a Hamming distance of 0.149 $\pm$ 0.062 and a BLEU score of 0.785 $\pm$ 0.082 for a bidirectional model with convolution, 81 embedding dimensions, and 32 hidden state dimensions. The highest cosine similarity (0.906 $\pm$ 0.087) was achieved by a 3-layer RNN model. We noted some differences in performance between models with and without convolution, as well as between unidirectional and bidirectional configurations, though these differences were not always substantial. Varying the embedding dimension (36, 81, 162) in RNN models resulted in similar performance levels. RNN models using center embedding showed different results compared to those using PxBy embedding. The performance of SSM and Attention models varied in comparison to RNN models across the different metrics, but no model type consistently outperformed the others across all measures.

\subsection{Identified Challenges}

Our initial results revealed several limitations in our embedding approach. While we observed variations in performance across different model configurations, the differences were often not substantial \cite{wang2020deep}. The PxBy embedding, which we initially considered promising, did not consistently outperform simpler approaches across all metrics and model types. Based on these findings, we recognized the need to refine our approach. The repetition of sequences in our generated output indicated that our embedding method might not be capturing the full range of patterns in our data \cite{bengio2013representation}.

\section{Optimizing Unified Representation}

\subsection{Refined Embedding Approach}

To address the challenges identified in our initial experiments, we propose a revised embedding strategy. Instead of using a convolutional approach, we now focus on six specific positions within each token, with the input dimension equal to the output dimension. This adjustment allows for a larger embedding size while potentially enhancing the model's ability to capture relevant patterns. We also recognized the need for a more flexible tokenizer \cite{kudo2018sentencepiece}. Our initial implementation proved cumbersome when generating new data, highlighting the importance of a more versatile approach. We are exploring methods to integrate all necessary functionality into the tokenizer itself, which should streamline our overall pipeline and potentially improve performance. These refinements reflect a shift from our initial exploration towards a more focused approach to unified representation. While our initial results provided valuable insights, they also revealed the complexities inherent in multimodal sequence modeling and the need for continuous iteration in our methods \cite{lecun2015deep}.

\subsubsection{Dataset Construction}

\sloppy

For our refined approach, we developed a new dataset combining images, text, and audio extracted from Pokemon sprite animations. This dataset, available on the Hugging Face Dataset Hub \cite{furfaro2024pixelbytes_project}, was compiled through web scraping of Pokepedia and includes descriptions of the Pokemon along with their associated cries. The dataset comprises animated, pixelated GIFs of Pokemon sprites as the visual component. The audio files are two-channel recordings: Channel 1 contains the original mono sound of the Pokemon cry, while Channel 2 features a filtered version simulating a bits Game Boy speaker output to verify our approach for control problems. This setup enables us to model a simplified dynamic physical system, where the original sound acts as the "action" input and the filtered output represents the "state" of the system. The transfer function of this bandpass filter can be approximated as:

\begin{equation}
H(s) = \frac{K \omega_n^2}{s^2 + 2\zeta\omega_n s + \omega_n^2}
\end{equation}

where $K$ is the gain, $\omega_n$ is the natural frequency, and $\zeta$ is the damping ratio. These parameters can be adjusted to closely match the frequency response of a Game Boy speaker.

\subsection{Enhanced Tokenization Strategy}

Building on our previous work, we developed an improved tokenization strategy using the ActionPixelBytesTokenizer. This tokenizer addresses multimodal data more effectively, including text, images, and audio, while maintaining a unified representation. It employs a combined vocabulary that includes ASCII bytes, RGB values from the NES palette, and action states for control and audio. This approach aims to create a consistent representation across different data types. For text processing, the tokenizer converts to lowercase ASCII bytes. Images are converted to Lab color space and quantized to the nearest NES palette color. Audio data is normalized and mapped to predefined action states, with the setpoint reset to zero (standard equilibrium). The token vocabulary now comprises 151 tokens, where index 0 corresponds to the null padding value, and indices 1 and 2 are transition values. A key aspect of the new tokenizer is its sequence construction method. Instead of using convolutional methods, we focus on six specific positions for each token to avoid repetition in the sequencing of a 3D zigzag scheme (Figure \ref{fig:approach_overview}). This approach creates context-target pairs that aim to capture relationships between neighboring tokens in both space and time. The sequence construction algorithm is detailed below:

\begin{algorithm}[ht]
\small
\caption{Create Sequence Data
\newline
\textbf{Input:} $\mathbf{X} \in \mathbb{R}^{T \times H \times W}$: context array
\newline
\textbf{Output:} $\mathbf{C} \in \mathbb{R}^{THW \times 6}$: context, $\mathbf{Y} \in \mathbb{R}^{THW \times 1}$: targets
\newline
\textbf{Note:} $T$: time steps, $H$: height, $W$: width
}
\begin{algorithmic}[0]
\State \textbf{Initialize:} $\mathbf{P} \in \mathbb{R}^{(T+1) \times (H+2) \times (W+2)}$ as padded array
\Function{CreateSequenceData}{$\mathbf{X}$}
    \State $\mathbf{P} \gets \text{Pad}(\mathbf{X}, (1,1,1,1,1,0), \text{mode='constant'}, \text{value=0})$
    \State $\mathbf{S}_1 \gets \mathbf{P}_{1:T, 1:H, 2:W+1}$
    \State $\mathbf{S}_2 \gets \mathbf{P}_{1:T, 2:H+1, 2:W+1}$
    \State $\mathbf{S}_3 \gets \mathbf{P}_{1:T, 3:H+2, 2:W+1}$
    \State $\mathbf{S}_4 \gets \mathbf{P}_{2:T+1, 1:H, 1:W}$
    \State $\mathbf{S}_5 \gets \mathbf{P}_{2:T+1, 2:H+1, 1:W}$
    \State $\mathbf{S}_6 \gets \mathbf{P}_{2:T+1, 1:H, 2:W+1}$
    \State $\mathbf{C} \gets \text{Stack}([\mathbf{S}_1, \mathbf{S}_2, \mathbf{S}_3, \mathbf{S}_4, \mathbf{S}_5, \mathbf{S}_6], \text{dim=-1})$
    \State $\mathbf{C} \gets \text{Reshape}(\mathbf{C}, [THW, 6])$
    \State $\mathbf{Y} \gets \text{Reshape}(\mathbf{X}, [THW, 1])$
    \For{$i \gets 2$ \textbf{to} $THW$}
        \State $\mathbf{C}_{i,6} \gets \mathbf{Y}_{i-1,1}$
    \EndFor
    \State \Return $\mathbf{C}, \mathbf{Y}$
\EndFunction
\end{algorithmic}
\end{algorithm}

This algorithm constructs a context array for sequence modeling. It pads the input to handle boundary conditions and extracts six slices to represent spatial and temporal relationships. The context array $\mathbf{C}$ is formed by stacking these slices, while the targets $\mathbf{Y}$ are derived from the input. The last column of $\mathbf{C}$ is set to the true previous value, implementing an autoregressive feature. The tokenizer is designed to handle various input combinations and includes functionalities such as special token handling and padding.

\subsection{Autoregressive Model Architecture}

Building upon our initial approach findings, we developed the aPxBySequenceModel architecture. This architecture is designed to handle both predictive and autoregressive tasks using a Long Short-Term Memory (LSTM) network. The model comprises three main components: an embedding layer, an LSTM sequence model, and a fully connected output layer. The embedding layer maps input tokens to a continuous vector space, with the embedding size calculated based on the input dimension. Specifically, the embedding size is determined by dividing the overall embedding size by the number of positions we focus on within each token. This approach aligns with our revised embedding strategy, which emphasizes six specific positions within each token (with padding 0 to avoid influencing training). The LSTM layer aims to capture temporal dependencies in the sequence data, potentially enhancing pattern recognition across different modalities. The model operates in two distinct modes:

\begin{itemize}
    \item \underline{Predictive mode:} In this configuration, the model takes six input values and attempts to predict only the next token.

    \item \underline{Autoregressive mode:} Here, the model's output dimension matches the input dimension (Figure \ref{fig:approach_overview}). Additionally, the output is restructured by multiplying it with the vocabulary size, enabling the model to generate sequences based on the learned representations.
\end{itemize}

During the forward pass, input data is processed through the embedding layer, then through the LSTM layers, and finally through the fully connected layer. The output shape is adjusted based on the operating mode, which may provide the flexibility we found lacking in our initial implementation.

\subsubsection{Model Training and Data Management}

For data management, we developed the \texttt{TokenPxByDataset} class to handle multimodal inputs, including text, image, and audio data. This class generates overlapping sequences from longer inputs, facilitating the model's capture of context across sequence boundaries. It optimizes memory usage through on-the-fly data retrieval, preparing samples only as needed. The class ensures consistent sequence lengths by implementing circular padding for sequences extending beyond an item's end. These features enable efficient processing of variable-length inputs during training.

Our training process incorporates several enhancements for efficiency and monitoring. In the \texttt{\_process\_epoch} function, we manage both autoregressive and non-autoregressive modes. For autoregressive mode, we reshape input and output sequences, using the input sequence as the target. In non-autoregressive mode, we flatten the outputs and use provided labels as the target. This flexibility allows the model to adapt to different tasks. The \texttt{train\_model} function alternates between training and validation phases, enabling regular performance evaluation. We employ gradient accumulation to simulate larger batch sizes, beneficial when GPU memory is limited. The training loop tracks both training and validation metrics (loss and accuracy) for each epoch, saving these metrics to a CSV file. We also implement model checkpointing to retain the best model based on validation loss.

\subsection{Performance Evaluation}

We evaluated three LSTM models: one in predictive mode and two in autoregressive mode, each with approximately 4 million parameters. The models were trained on Kaggle using T4 GPUs. We utilized an embedding size of 128, a hidden size of 512, and two layers. Training was conducted for 100 epochs with a batch size of 32, a learning rate of 0.001, and a sequence length of 1024.

\subsubsection{Results Comparison}

To manage data proportions, we applied different reduction strategies for image and audio data. This was particularly important for audio data in the autoregressive mode, as it contains more null values to predict, which could potentially impact training. Table \ref{tab:model_comparison_lstm} presents the final performance metrics after training.

\begin{table}[ht]
\centering
\caption{Comparison of model performance after 100 epochs}
\label{tab:model_comparison_lstm}
\begin{tabular}{|l|c|c|c|c|}
\hline
\textbf{Model} & \textbf{Train Loss} & \textbf{Train Accuracy} & \textbf{Val Loss} & \textbf{Val Accuracy} \\
\hline
Autoregressive (2,2) & 0.2211 & 0.9329 & 0.4519 & 0.8852 \\
Autoregressive (4,2) & 0.2346 & 0.9290 & 0.4914 & 0.8726 \\
Predictive (2,2) & 0.8810 & 0.7440 & 2.1144 & 0.6009 \\
\hline
\end{tabular}
\end{table}

The numbers in parentheses indicate the reduction factors for (audio, image) data. Lower loss and higher accuracy indicate better performance. Both autoregressive models outperformed the predictive model in terms of accuracy and loss on both training and validation sets. The autoregressive model with balanced reduction (2,2) achieved slightly better results than the one with more aggressive audio reduction (4,2). Although the predictive model had lower performance overall, it still reached reasonable accuracy given the task complexity. This suggests that our pixelbyte representation is effective, especially when using autoregressive learning.

\begin{figure}[ht]
\centering
\includegraphics[width=\textwidth]{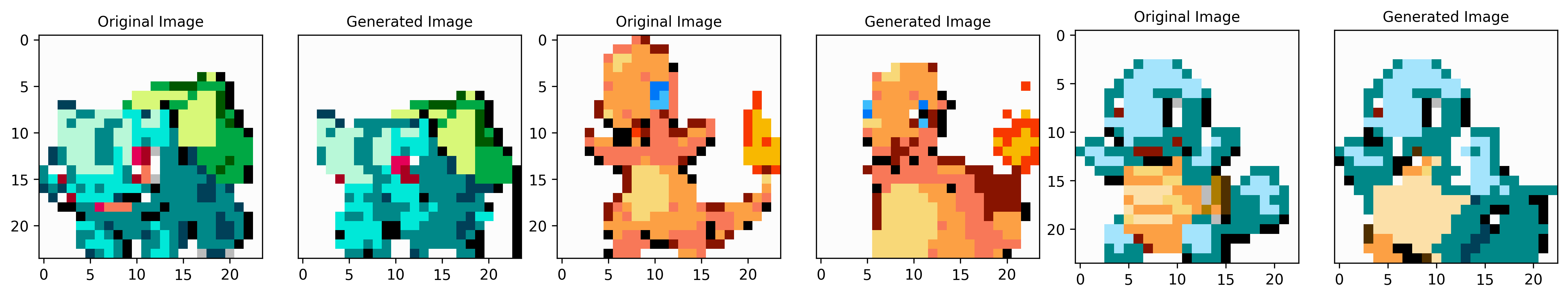}
\caption{Generation results for the 1st Generation Starter Pokémon using the autoregressive model with a temperature of 0.1. Reference images and generated images are paired sequentially.}
\label{fig:generation_results}
\end{figure}

The complete frame generation results, as shown in Figure \ref{fig:generation_results}, demonstrate spatial consistency in image creation, despite the model being unidirectional. This suggests that data representation may be more important than the architecture itself, aligning with recent findings in multimodal learning \cite{baltruvsaitis2018multimodal}. The generated images show coherence in colors and sizes, indicating that the model has captured some key visual features. While our model does not match the image quality of state-of-the-art diffusion models \cite{dhariwal2021diffusion}, it demonstrates the capability to capture multiple modalities (image, text, audio) within a single framework. This aligns with recent trends in multimodal co-learning \cite{liang2022mind}, although further optimization is needed to improve generation quality. The trained models are available on the Hugging Face Model Hub \cite{furfaro2024pixelbytes_project} for further examination and replication of our results.

\section{Toward application in control problem}

The auto-regressive approaches we previously explored showed promising results in image sprite generation. However, these structures may have limitations for rapid image generation and training of bidirectional models. To address these challenges, we incorporated diffusion-based models into our research, which offer more flexibility for various applications. Diffusion processes have shown good results in control optimization problems, as seen in recent work on planning with diffusion \cite{janner2022planning}. In this part, we compare a semi-diffusive model with an auto-regressive model in the context of optimal control.

\subsection{Dataset Optimal Control Construction}

To create our optimal control dataset, we used linear systems that are stable, observable, and controllable, based on Kalman's principles \cite{kalman1960contributions}. We followed these steps:

\begin{enumerate}
\item We define the linear system using state-space matrices (A, B, C, D) as described in control theory \cite{ogata2009modern}:

   \begin{equation}
   \begin{aligned}
   \dot{x} &= Ax + Bu \\
   y &= Cx + Du
   \end{aligned}
   \end{equation}

   where $x$ is the state vector, $u$ is the input vector, and $y$ is the output vector.

\item We check that the system meets Kalman's conditions:
   \begin{itemize}
   \item Stability: For a transfer function $H(s) = \frac{N(s)}{D(s)}$, all eigenvalues $\lambda_i$ of A and all roots $p_i$ of $D(s)$ must satisfy:
     \begin{equation}
     \begin{cases}
       Re(\lambda_i) < 0 \text{ and } Re(p_i) < 0 & \text{for continuous-time systems} \\
       |\lambda_i| < 1 \text{ and } |p_i| < 1 & \text{for discrete-time systems}
     \end{cases}
     \end{equation}
   \item Controllability: rank$[B \quad AB \quad A^2B \quad ... \quad A^{n-1}B] = n$, where $n$ is the number of states.
   \item Observability: rank$[C \quad CA \quad CA^2 \quad ... \quad CA^{n-1}]^T = n$
   \end{itemize}

\item For each (initial condition, setpoint) pair, we calculate the optimal control sequence using the Linear Quadratic Gaussian (LQG) approach \cite{athans1972determination}. The LQG controller minimizes:

   \begin{equation}
   J = \int_0^\infty (x^TQx + u^TRu) dt
   \end{equation}

   where $Q$ and $R$ are weighting matrices.
\end{enumerate}

We also added bang-bang controller responses with different time delays and noise levels. The bang-bang control law is:

\begin{equation}
u(t) = 
\begin{cases} 
u_{max}, & \text{if } e(t) > 0 \\
u_{min}, & \text{if } e(t) < 0
\end{cases}
\end{equation}

where $e(t)$ is the error signal.

Lastly, we sampled and quantized the continuous signals to create a discrete representation for our model. The dataset is available on Hugging Face \cite{furfaro2024pixelbytes_project}.

Table \ref{tab:param_ranges} shows the ranges of parameters used in dataset generation:

\begin{table}[ht]
\caption{Parameter Ranges for Dataset Generation}
\label{tab:param_ranges}
\centering
\begin{tabular}{|l|c|}
\hline
\textbf{Parameter} & \textbf{Range} \\
\hline
System order & 1 - 5 \\
Eigenvalues of A & [-5, -0.1] \\
Elements of B & [-1, 1] \\
Elements of C & [0, 1] \\
Initial conditions & [-1, 1] \\
Setpoints & [-1, 1] \\
Bang-bang delay & 0 - 5 time steps \\
Noise standard deviation & 0 - 0.1 \\
\hline
\end{tabular}
\end{table}

\subsection{Diffusion Model Architecture}

We use a modified bidirectional LSTM to handle discretized control sequences. Our model, shown in Algorithms 3 and 4, can be used in various generation and control scenarios. The model can switch between predictive and diffusion modes, which may be useful for different types of training problems.


\begin{multicols}{2}
\begin{algorithm}[H]
\caption{Forward Pass (F)
\newline
\textbf{Input:} $x$: input, $t$: time step, $m$: mask
\newline
\textbf{Output:} processed output
\newline
\textbf{Note:} E: Embedding, S: SequenceModel, 
\newline
FC: FullyConnected
\newline
$n_d$: num\_diffusion\_steps, $s_l$: seq\_len, 
\newline
$p_d$: pxby\_dim, $p_e$: pxby\_emb, $b_s$: batch\_size
}
\small
\begin{algorithmic}
\Procedure{F}{$x, t, m$}
    $x \gets E(x)$

    \If{$o = \text{"d"}$}
        \If{$t = \text{null}$} 
            $t \gets \text{R}(n_d - 1, n_d + 1)$
        \EndIf

        \If{$m = \text{null}$}
            $p \gets \text{R}(0, s_l, 3s_l/4)$

            $m \gets \text{CM}(p, s_l, p_d, p_e)$
        \EndIf

        $\alpha \gets 1 - t / n_d$

        $n \gets \text{RN}(x.\text{shape})$

        $x \gets \text{W}(m = 1, x, (1 - \alpha) * n + \alpha * x)$
    \EndIf

    $x \gets S(x)$

    $x \gets \text{FC}(x)$

    \If{$o = \text{"p"}$}
        \Return $x$
    \Else
        \Return $\text{RS}(x, [b_s, s_l, p_d, -1])$
    \EndIf
\EndProcedure
\end{algorithmic}
\end{algorithm}

\columnbreak

\begin{algorithm}[H]
\caption{Generation Process (G)
\newline
\textbf{Input:} $i$: input, $T$: temperature, $m_l$: max length
\newline
\textbf{Output:} generated sequence
\newline
\textbf{Note:} F: Forward, S: Softmax, M: Multinomial
\newline
R: RandomInt, RN: RandomNormal, W: Where
\newline
CM: CreateMask, RS: Reshape, SU: ScatterUpdate
}

\small
\begin{algorithmic}
\Procedure{G}{$i, T, m_l$}
    $c \gets i$

    \For{$j \gets 1$ to $m_l$}
        $o \gets F(c)$

        \If{$ob = \text{"p"}$}
            $p \gets S(o[:, -1] / T)$

            $n \gets M(p)$

            $c[:, -(j+1)] \gets n$
        
        \ElsIf{$ob = \text{"d"}$}
            $ps \gets \text{RP}()$

            $p \gets S(o[:, ps] / T)$

            $t \gets M(p)$

            $c \gets SU(c, ps, t)$

        \Else
            $p \gets S(o[:, -1] / T)$

            $n \gets M(p)$

            $c \gets C(c, n)$
        \EndIf
        \newline

    \EndFor
    \newline

    \Return $c$

\EndProcedure
\end{algorithmic}
\end{algorithm}
\end{multicols}

\subsubsection{Model training and performance}

We trained the models on Kaggle using T4 GPUs, with a batch size of 32, sequence length of 512 (256 overlap), 256 hidden sizes for unidirectional and 128 for bidirectional diffusion model, and learning rate of 0.001 for 200 epochs. The difference in size between the bidirectional and unidirectional versions is to avoid doubling the network weight size. During training, we faced challenges with our 6-input dataset's unbalanced nature. The '0' token is frequent and, while not in the embedding, affects the diffusion process and output generation. To address this, we used a weight normalization for the optimizer based on token frequency:

\begin{equation}
    w_i' = w_i \cdot \frac{\sqrt{\sum_{j} f_j}}{\sqrt{f_i}}
\end{equation}

where $w_i'$ is the normalized weight, $w_i$ is the original weight, and $f_i$ is the frequency of token $i$ in the dataset. We tested different model sizes to balance performance and efficiency. Table \ref{tab:model_sizes} shows our train and evaluation results:

\begin{table}[ht]
\centering
\begin{tabular}{lcccc}
\hline
Model Size & Direction & Train Type & Loss & Accuracy (\%) \\
\hline
1.2M & Unidirectional & Auto-regressive & 0.19 & 92.1 \\
0.8M & Bidirectional & Diffusion & 0.30 & 95.9 \\
\hline
\end{tabular}
\caption{Diffusion Model Architectures and Performance}
\label{tab:model_sizes}
\end{table}

Our results show that the bidirectional diffusion model, with fewer parameters, achieves higher accuracy than the unidirectional auto-regressive model. This suggests the diffusion-based approach may better capture control sequence complexities. The trained models are available on the Hugging Face Model Hub \cite{furfaro2024pixelbytes_project}.

\subsection{Setpoint following with diffusion generation}

We tested our training on a simple case: a transfer function of \( \frac{1}{s+1} \) with a setpoint of 0.5. We used the "Gymsetpoint" library \cite{furfaro2024gymsetpoint}, made for reinforcement learning of an LTI system with input setpoint and targeted output action. We start with a noisy bang-bang controller. After 126 time steps, we begin generating the action to reach our setpoint step-by-step. We apply a mask to \( N \) repeated action values (targets) and an incomplete mask on the expected state setpoints. We then use the average of the targets generated by diffusion for the next action step only. We generate 125 consecutive actions to reach 251 total steps. We did this experiment 100 times.

\begin{figure}[ht]
\centering
\includegraphics[width=\textwidth]{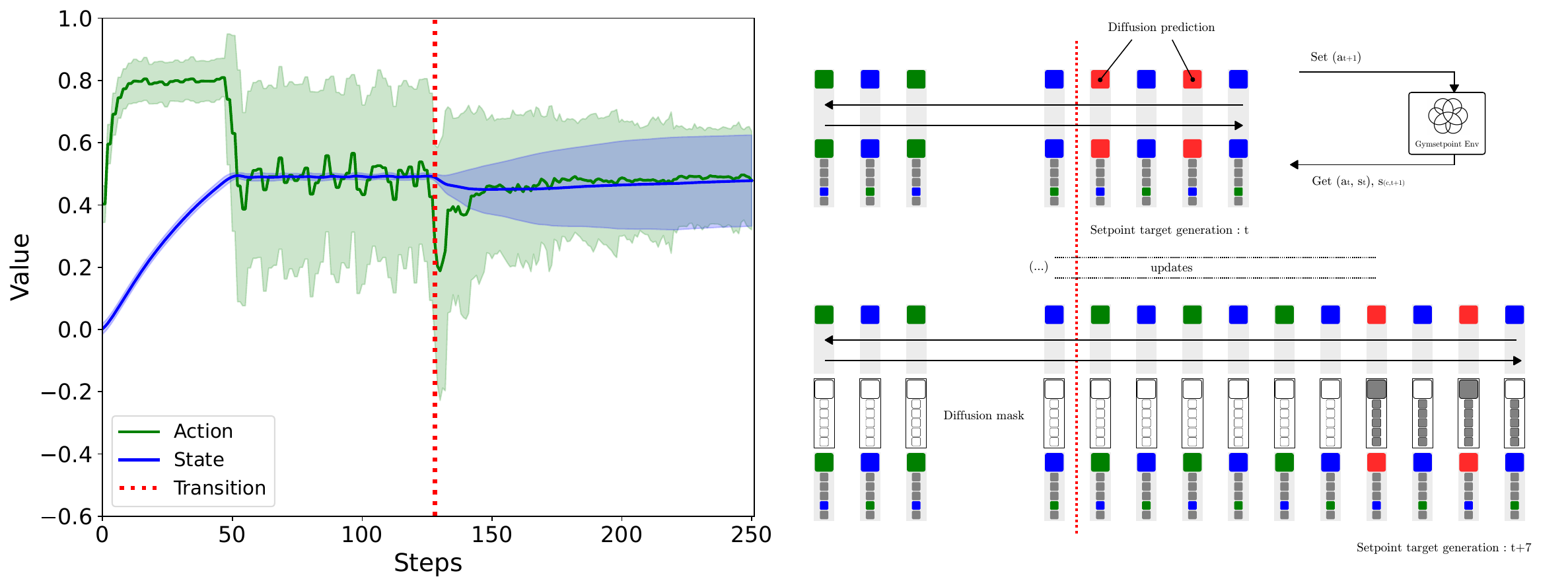}
\caption{Generation approach for control problem. Left: generation result for linear problem. Right: generation step with setpoint and targeted diffusion action following input mask.}
\label{fig:control_approach_results}
\end{figure}

Figure \ref{fig:control_approach_results} shows that the diffusion-based control is more precise and robust than bang-bang control, despite more input variability. The bang-bang control (left part of the figure) has high input variability before the diffusion-controlled region. The diffusion control works well but can sometimes produce odd results. This means we might need extra checks for real-world use. These results suggest our diffusion model could perform better than some control methods in certain cases. However, the occasional odd results show we need good error-checking and backup plans in real use, or we might need to retrain the model for specific cases.

\section{Discussion and Future Directions}

Our experiments with various model architectures for unified text and image generation have given us new insights and led to changes in our approach. We initially explored bidirectional RNN models using PixelBytes (PxBy) embedding with convolutional layers, expecting better multimodal data representation. However, we found limitations in this approach, leading us to reconsider our tokenizer design.

Our results with LSTM models have been informative. The autoregressive models performed significantly better than the predictive model, suggesting that keeping equal input and output dimensions is important for our task. This aligns with recent research on preserving structural information in multimodal embeddings \cite{verHo2021efficient}. The performance difference between the two autoregressive models highlights the impact of data balancing strategies. The model with balanced reduction (2,2) for audio and image data showed slightly better results, indicating potential overfitting with animated images. We now aim for a more versatile solution that can handle all aspects of data preparation and support true autoregressive modeling. Our \texttt{TokenPxByDataset} class remains valuable, but we are working to integrate its functionality more closely with our revised tokenizer for a more streamlined data pipeline.

While we explored both RNN and State Space Models (SSM), with SSMs showing promising rapid convergence \cite{dao2024transformers}, we're now focusing on simplifying the overall architecture. Our approach offers an alternative to models like ImageBind \cite{girdhar2023imagebind}, as our results suggest it may be possible to unify modalities without relying on intermediate representations. We now see potential benefits in allowing emergent properties to develop within a simplified framework. Moving forward, we plan to refine our strategy to better use specific input positions for high-definition image and sound by including basic diffusion. Our exploration of diffusion-based models for setpoint control tasks has shown interesting results in terms of generalization, expanding the potential applications of our multimodal approach to control problems. This work also opens new possibilities for applying approaches like Diffusion-LM \cite{li2022diffusion} and Autoregressive Diffusion Models \cite{hoogeboom2021autoregressive}, which may offer new perspectives on multimodal sequence modeling. While our work is ongoing, it aligns with recent trends in multimodal AI research \cite{baltruvsaitis2018multimodal} and could potentially provide a versatile foundation for various multimodal tasks.

\bibliographystyle{plain}
\bibliography{references}

\end{document}